\documentclass{article}

\usepackage{amssymb}
\usepackage{microtype}
\usepackage{graphicx}
\usepackage{wrapfig}
\usepackage{subfigure}
\usepackage{booktabs} 
\usepackage{natbib}
\usepackage{booktabs,multirow}
\usepackage[table, svgnames, dvipsnames]{xcolor}
\usepackage{makecell, cellspace, caption}

\usepackage[framemethod=TikZ]{mdframed}
\usepackage{hyperref}
\usepackage{listings}
\usepackage{color}

\definecolor{dkgreen}{rgb}{0,0.6,0}
\definecolor{gray}{rgb}{0.5,0.5,0.5}
\definecolor{mauve}{rgb}{0.58,0,0.82}

\usepackage{empheq}
\usepackage{xcolor}

\usepackage{hyperref}

\usepackage[accepted]{icml2023}

\usepackage{amsmath}
\usepackage{amssymb}
\usepackage{mathtools}
\usepackage{amsthm}

\usepackage[capitalize,noabbrev]{cleveref}

\theoremstyle{plain}
\newtheorem{theorem}{Theorem}[section]

\newtheorem{lemma}[theorem]{Lemma}

\theoremstyle{definition}
\newtheorem{definition}[theorem]{Definition}

\theoremstyle{remark}

\usepackage[textsize=tiny]{todonotes}

\icmltitlerunning{Global Selection of Contrastive Batches via Optimization on Sample Permutations}

\begin{document}

\twocolumn[
\icmltitle{Global Selection of Contrastive Batches via Optimization on Sample Permutations}



\icmlsetsymbol{equal}{*}

\begin{icmlauthorlist}
\icmlauthor{Vin Sachidananda}{stanford}
\icmlauthor{Ziyi Yang}{msft}
\icmlauthor{Chenguang Zhu}{msft}
\end{icmlauthorlist}

\icmlaffiliation{stanford}{Stanford University and Two Sigma Ventures}
\icmlaffiliation{msft}{Knowledge and Language Team, Azure Cognitive Services Research, Microsoft Research, Redmond, WA}

\icmlcorrespondingauthor{Vin Sachidananda}{vinsachida@gmail.com}

\icmlkeywords{Contrastive Learning, Representation Learning, Optimization, Language Models}

\vskip 0.3in
]



\printAffiliationsAndNotice{}  

\begin{abstract}
Contrastive Learning has recently achieved state-of-the-art performance in a wide range of unimodal and multimodal tasks. Many contrastive learning approaches use mined hard negatives to make batches more informative during training but these approaches are inefficient as they increase epoch length proportional to the number of mined negatives and require frequent updates of nearest neighbor indices or mining from recent batches. In this work, we provide an alternative to hard negative mining, Global Contrastive Batch Sampling (GCBS), an efficient approximation to the batch assignment problem that upper bounds the gap between the global and training losses, $\mathcal{L}^{Global} - \mathcal{L}^{Train}$, in contrastive learning settings. Through experimentation we find GCBS improves state-of-the-art performance in sentence embedding and code-search tasks. Additionally, GCBS is easy to implement as it requires only a few additional lines of code, does not maintain external data structures such as nearest neighbor indices, is more computationally efficient than the most minimal hard negative mining approaches, and makes no changes to the model being trained. Code is available at \href{https://github.com/vinayak1/GCBS}{https://github.com/vinayak1/GCBS}. 
\end{abstract}

\section{Introduction}
\label{Introduction}

Contrastive Learning is used ubiquitously in training large representation models, such as transformers, and has been shown to achieve state-of-the-art performance in a wide range of unimodal and multimodal tasks across language, vision, code, and audio \cite{simclr, gao2021simcse, jiang2022promcse, guo-etal-2022-unixcoder, coca, clip, dalle, speech-cl, yang2021universal}. In supervised contrastive learning, one is given a paired dataset $(X, Y)$ each with $N$ samples, where $x_i \sim y_i$ such as similar sentences, code and corresponding language descriptors, or images and their captions. For unsupervised settings, $X, Y$ are alternative "views" of the same sample often constructed through data augmentation schemes or independent dropout masks \cite{simclr, gao2021simcse}. Then batches of rows, $B$, are sampled from this pair of datasets and a model $f(\cdot)$ is trained to concurrently maximize inner products for outputs of similar (positive) data inputs, $f(x_i)^T f(y_i)$, and minimize inner product for outputs of dissimilar (negative) data inputs $f(x_i)^T f(y_j), i,j \in B, i \neq j$.

Due to batch size constraints from hardware limitations, for a fixed batch size $k$, only $Nk$ inner products of the total $N^2$ in $f(X)f(Y)^T$ are observed in the training loss for each epoch of training. Through the rest of this paper, we will refer to this observed training loss over $Nk$ inner products as $\mathcal{L}^{Train}$ and the total loss over $N^2$ inner products as $\mathcal{L}^{Global}$. It has been observed, both in contrastive metric and representation learning \cite{saunshi, hal_manifold, Xuan2020HardNE, mishchuk, wu_iccv, songCVPR16, schroff, harwood, Ge2018DeepML}, that in order for batches to be informative during training, they should be constructed to contain "hard-negatives", or large values of $f(x_i)^T f(y_j), i \neq j$. Additionally, it has been shown that including hard negatives in batches better approximates global losses \cite{zhang-stratos-2021-understanding}.

Currently, approaches for constructing batches, and controlling $\textit{which}$ inner products of the total $N^2$ should be used for training, broadly fall into one of two categories. One either uses random sampling or mines nearest neighbors of the reference sample $x_i$ in order to greedily insert hard negatives into the same batch as $x_i$. While greedily inserting hard negatives is effective in practice \cite{mixup, xiong2021approximate}, these methods incur large costs both in time and resources as mining $l < k$ hard negatives per reference sample increases each training epoch by a factor $l$ and often requires maintaining and reranking nearest neighbor indices on expensive accelerated hardware during training. For instance, if $5$ hard negatives from $Y$ are mined for each sample in $X$ during batch construction one will increase the training time of a single epoch by a factor $5$, not including time taken for constructing nearest neighbor indices. 

\begin{figure*}[ht]
\centering
\includegraphics[width=\textwidth]{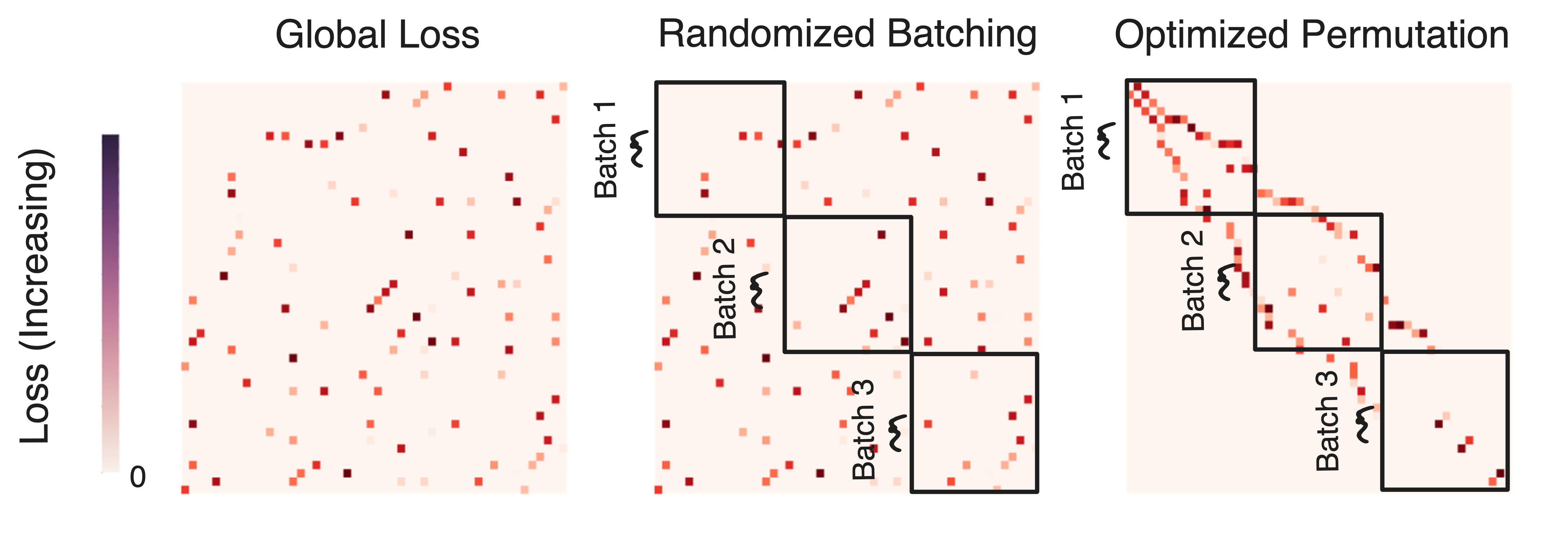}
\caption{Visualization of inner products of $f(X)f(Y)^T$ in global, training with random sampling, and training with permutation optimized sampling for contrastive losses.}
\label{GCBS-visual}
\end{figure*}

Furthermore, hard negative mining often requires frequent reranking to prevent negative anchors from being sampled from stale nearest neighbor indices. Work on momentum based memory banks have found that hard negative mining is especially useful with small lookback intervals (i.e. 2-4 previous batches) \cite{Wang2021CrossBatchNS}. In this paper, we propose a global alternative to hard negative mining, Global Contrastive Batch Sampling (GCBS), which seeks to efficiently learn a permutation over samples in $X$ and $Y$ to increase the likelihood of hard negatives before each epoch rather than through greedy insertion during training. In Figure \ref{GCBS-visual} above, we visually depict $\mathcal{L}^{Global}$ and $\mathcal{L}^{Train}$ along with the modifications on batches, and therefore the observed loss, for our proposed approach GCBS.

First, we show theoretically that the upper bound on $\mathcal{L}^{Global} - \mathcal{L}^{Train}$, with no oversampling or assumptions on the data/model for commonly used scaled cross entropy losses, such as NT-Xent \cite{ntxent}, are only dependent on batch assignments, total samples $N$, and batch size $k$. We prove that, for fixed $N, k$, this upper bound is minimized as a Quadratic Bottleneck Assignment Problem which seeks to maximize the number of hard negatives in batches by learning a permutation $\pi \in \Pi_N$ on the rows of $X$ and $Y$. We then formulate an $\mathcal{\tilde{O}}(N^2)$ approximation for optimizing over this permutation, GCBS, and show that it is more efficient than any hard negative mining approaches, even for $l=1$, per training epoch. We analyze the loss behavior of GCBS and show that GCBS better approximates the total contrastive loss. Lastly, we empirically evaluate GCBS in the context of supervised contrastive finetuning for sentence embedding (STS) and code search (CosQA, AdvTest, CodeSearchNet) and achieve state-of-the-art performance for all of these tasks.

In this work, we summarize our contributions as follows:
\begin{enumerate}
\itemsep0em 
  \item We prove that the upper bound of the gap between the total and observed losses in contrastive learning for a fixed batch size $B$ without oversampling, $\mathcal{L}^{Global} - \mathcal{L}^{Train}$, is constrained by a Quadratic Bottleneck Assignment Problem and can be relaxed to a Matrix Bandwidth Minimization problem.
  \item We formulate a $\mathcal{\tilde{O}}(N^2)$ time and $\mathcal{O}(Nk)$ space complexity approximation to the Matrix Bandwidth Minimization problem, GCBS, using the Cuthill-Mckee heuristic and implement this algorithm in less than 50 lines of PyTorch.  
  \item We analyze the loss behavior of GCBS and show that, in sentence embedding and code-search tasks, GCBS better approximates the total contrastive loss.  
  \item We empirically evaluate GCBS and achieve state-of-the-art performance on the STS taskset for sentence embeddings. Additionally, we achieve state-of-the-art performance for the CosQA, AdvTest, and CodeSearchNet tasks for joint programming language-natural language embeddings.
\end{enumerate}

The rest of this paper is organized as follows. In Section \ref{Related-work} we discuss related work. In Section \ref{Loss-bounds}, we derive upper bounds on the gap between total and observed losses and in Section \ref{bound_qbap} formulate these bounds as Quadratic Assignment Problems. In Section \ref{sec_approx_qbap}, we relax our QBAP to a Matrix Bandwidth Minimization problem, introduce our proposed method, GCBS, for approximating global contrastive losses, and provide implementation details. In Section \ref{Exps}, we provide experimental results for sentence embedding and code search tasks using GCBS. Section \ref{Discussion} provides discussion and Section \ref{Conclusion} concludes the paper.

\section{Related Work}
\label{Related-work}
\subsection{Contrastive Representation Learning}
Contrastive Learning has been used ubiquitously for vision, language, and audio representation learning. In vision tasks, SimCLR \cite{simclr} showed that using augmented views of the same image as positive samples and the NT-Xent objective \cite{ntxent} improves performance of unsupervised classification. MoCo \cite{moco_v1, moco_v2} used memory banks of negative samples from recent batches to increase the effective contrastive batch size, and \citet{sylvain, khosla} show improvements using supervised contrastive frameworks. For sentence embedding tasks, contrastive learning has been used both in pretraining \cite{logeswaran2018an}, finetuning and continuous prompt tuning settings \cite{gao2021simcse, jiang2022promcse} to provide state-of-the-art performance. Additionally, contrastive learning has been used extensively to align representations across different modalities for downstream use in multimodal tasks such as those involving language/code, language/vision, and vision/decision making \cite{guo-etal-2022-unixcoder, feng-etal-2020-codebert, graphcodebert, clip, dalle, laskin_srinivas2020curl}.

\subsection{Hard Negative Mining in Metric and Contrastive Learning}
Selection of hard negatives during batch construction is well-studied and has been shown, both theoretically and empirically, to improve metric and contrastive learning \cite{saunshi, hal_manifold, Xuan2020HardNE, mishchuk, wu_iccv}. Prior work in metric learning \cite{songCVPR16, schroff, harwood, Ge2018DeepML} has observed that "hard negatives", or negatives which are difficult to discriminate against with respect to a particular query's embedding, are beneficial for downstream classifier performance. In contrastive learning, \citet{mixup} uses Mixup to generate hard negatives in latent space. \citet{chuang} proposes a debiased contrastive loss which approximates the underlying “true” distribution of negative examples and \citet{mutual_contrast} studies the effect of restricting negative sampling to regions around the query using a variational extension to the InfoNCE objective. In \citet{adversarial_1, adversarial_2} adversarial examples are used to produce more challenging positives and hard negatives. In \citet{xiong2021approximate}, nearest neighbor indices and a secondary model from prior checkpoints are used to mine hard negatives for text retrieval tasks.

\citet{robinson2021contrastive} reweights negative samples based on their Euclidean distance and debiases positive samples in order to control the level of difficulty in unsupervised contrastive learning. \citet{kalantidis} show that harder negative examples are needed to improve performance and training speed in vision tasks and propose adding "synthetic" hard negatives in training batches using convex combinations of nearest neighbors.

\subsection{Quadratic Assignment Problems}

The Quadratic Assignment Problem (QAP), stemming from facilities locations problems \cite{koopmans}, in combinatorial optimization seeks to minimize the total cost of assigning $n$ facilities to $n$ locations. Formally, one seeks to optimize $\min _{\pi\in \Pi _{n}}\operatorname {Tr} (W\pi D \pi^{T})$ over $\Pi _{n}$, the set of $n \times n$ permutation matrices, for a given cost matrix $W \in \mathbb{R}^{n \times n}$ and distance matrix $D \in \mathbb{R}^{n \times n}$. The Quadratic Bottleneck Assignment Problem (QBAP) \cite{steinberg} takes a similar form but minimizes the maximum cost rather than the total cost, $\min _{\pi \in \Pi _{n}} \max_{i,j} (W \pi D \pi^{T})_{i,j}$. The Graph Bandwidth Minimization Problem, seeks to minimize the dispersion of nonzero costs from the main diagonal for a sparse distance matrix $D$ and is a special case of QBAP in which the cost matrix $W$ increases monotonically in $|i-j|$. In this paper, we prove that minimizing the upper bound between the total and the observed training losses $\mathcal{L}^{Global} - \mathcal{L}^{Train}$ over a pair of datasets $X, Y$ is bounded by Quadratic Assignment Problems and approximated by a Graph Bandwidth Minimization Problem. This connection is shown visually in Figure \ref{GCBS-visual}. As all of the aforementioned problems are NP-Hard, we utilize the Cuthill-McKee algorithm \cite{cuthill-mckee}, a $\mathcal{\tilde{O}}(N^2)$ approximation for bandwidth minimization.

\section{Global and Training Losses for Cross-Entropy based Contrastive Objectives}
\label{Loss-bounds}

In this section, we characterize the gap between training and global losses in supervised contrastive learning for the Normalized Temperature-scaled Cross Entropy (NT-Xent) loss \cite{ntxent}. The NT-Xent loss is a ubiquitous contrastive objective used in state-of-the-art models for sentence embedding, code-language tasks and vision-language tasks \cite{gao2021simcse, jiang2022promcse, guo-etal-2022-unixcoder, simclr, clip}.

Let $X, Y \in \mathbb{R}^{n \times d}$ be the output representations for two paired datasets each with $N$ samples. Consider a supervised setting where $x_i, y_i$ are considered "positive" pairs and $x_i, y_j$ are considered negative pairs $\forall i \neq j$. Note that the analysis we provide can be modified to incorporate multiple positives, such as those from class information, which will have tighter bounds in terms of the number of samples $N$ and the batch size $k$. Additionally, let $\tau \in \mathbb{R}_+$ be a tunable temperature parameter which scales logit values in the objective. When $\tau$ is small, the NT-Xent loss is a proxy for the hard maximum loss. Assume that all representations have been normalized, that is $\| x_i\|_2, \| y_i\|_2,  = 1$ $\forall i$.

\subsection{The Global and Training NT-Xent objectives, $\mathcal{L}^{Global}$ and  $\mathcal{L}^{Train}$}
First, we provide the contrastive loss over all $N^2$ pairs of inner products between $X$ and $Y$. We call this the global objective as it contains all pairwise contrastive information and, in the absence of resource constraints, is the loss one would seek to minimize. The Global NT-Xent objective is given as follows:
\begin{definition}[Global NT-Xent objective]
\textit{The Global NT-Xent objective is given as follows:}
$\begin{aligned}
\mathcal{L}^{Global} &= -\frac{1}{N} \sum_{i=1}^{N} \log \frac{\exp( x_i^T y_i{\tau}^{-1})}{\sum_{\substack{j=1}}^N \exp(x_i^T y_j{\tau}^{-1})}. \\
&= \frac{1}{N} \sum_{i=1}^{N} \bigl(-x_i^T y_i \tau^{-1} + \log \sum_{\substack{j=1}}^N \exp(x_i^T y_j {\tau}^{-1})\bigr).
\end{aligned}$

\label{defn:global-ntxent}
\end{definition}

Due to memory constraints, during training one does not make all $N^2$ comparisons over pairs in $X$ and $Y$ during a training epoch. Instead, each sample $x_i$ is only contrasted against $k$ in-batch samples in $Y$, its positive anchor $y_i$ and $k-1$ negative anchors. This observed training loss will be strictly less than the global loss as it makes $k$ comparisons out of $N$ total for each sample. For a fixed batch assignment $B$, let $B_i$ be the indices of rows in $Y$ contained in a batch with $x_i$. The training NT-Xent objective is given as follows: 

\begin{definition}[Training NT-Xent objective]
\textit{The Training NT-Xent objective is given as follows:}
$\begin{aligned}
\mathcal{L}^{Train} &= -\frac{1}{N} \sum_{i=1}^{N} \log \frac{\exp( x_i^T y_i{\tau}^{-1})}{\sum_{\substack{j \in B_i}} \exp(x_i^T y_j{\tau}^{-1})}.\\
&= \frac{1}{N} \sum_{i=1}^{N} \bigl(-x_i^T y_i \tau^{-1} + \log \sum_{\substack{j\in B_i}} \exp(x_i^T y_j {\tau}^{-1})\bigr).
\end{aligned}$

\label{defn:training-ntxent}
\end{definition}

\subsection{Minimizing the gap between $\mathcal{L}^{Global}$ and $\mathcal{L}^{Train}$}
For a fixed set of batches $B$, we will first provide upper bounds on $\mathcal{L}^{Global}$ and lower bounds on $\mathcal{L}^{Train}$ using Log-Sum-Exp properties \cite{OptModels}. Using the upper bound for Log-Sum-Exp, the following bound on $\mathcal{L}^{Global}$ can be obtained where equivalence is attained when all inner products have the same value. 

\subsubsection{Upper bound on $\mathcal{L}^{Global}$}

\begin{lemma}[Upper bound on $\mathcal{L}^{Global}$]
With Log-Sum-Exp properties \cite{OptModels}, $\mathcal{L}^{Global}$ with the NT-Xent contrastive objective can be upper bounded as:
$\begin{aligned} \mathcal{L}^{Global} &=\frac{1}{N} \sum_{i=1}^{N} -x_i^T y_i \tau^{-1} + \log \sum_{\substack{j=1}}^N \exp(x_i^T y_j {\tau}^{-1}) \\
&\leq \frac{1}{N} \sum_{i=1}^{N} -x_i^T y_i \tau^{-1} + \log (N \max_j \exp(x_i^T y_j {\tau}^{-1}) \\
&= \frac{1}{N} \sum_{i=1}^{N} {\tau}^{-1}(-x_i^T y_i {\tau}^{-1} + \max_{j} x_i^T y_j) + \log N.
\end{aligned}$
\label{lem:ub-global}
\end{lemma}

\subsubsection{Lower Bounds on $\mathcal{L}^{Train}$}

Two lower bounds can be derived for $\mathcal{L}^{Train}$, first using the translation identity property \cite{Nielsen-Sun} and then using the standard lower bound for Log-Sum-Exp \cite{OptModels}.

\begin{lemma}[First lower bound on $\mathcal{L}^{Train}$ using Translation Identity]
With the Log-Sum-Exp translation identity property \cite{Nielsen-Sun}, $\mathcal{L}^{Train}$ with the NT-Xent contrastive objective can be bounded as:
$\begin{aligned}
\mathcal{L}^{Train} &= \frac{1}{N} \sum_{i=1}^{N} -x_i^T y_i \tau^{-1} + \log \sum_{\substack{j\in B_i}} \exp(x_i^T y_j {\tau}^{-1}) \\
&\geq \frac{1}{N} \sum_{i=1}^{N} -x_i^T y_i \tau^{-1} + \log (k \min_{j\in B_i} \exp(x_i^T y_j {\tau}^{-1})) \\
&= \frac{1}{N} \sum_{i=1}^{N} {\tau}^{-1}(-x_i^T y_i + \min_{j\in B_i} x_i^T y_j) + \log k.
\end{aligned}$
\label{lem:lb-global-1}
\end{lemma}

\begin{lemma}[Second lower bound on $\mathcal{L}^{Train}$ using standard Log-Sum-Exp bound]
With Log-Sum-Exp properties \cite{OptModels}, $\mathcal{L}^{Train}$ with the NT-Xent contrastive objective can be bounded as:

$\begin{aligned}
\mathcal{L}^{Train} &= \frac{1}{N} \sum_{i=1}^{N} -x_i^T y_i \tau^{-1} + \log \sum_{\substack{j\in B_i}} \exp(x_i^T y_j {\tau}^{-1}) \\
&\geq \frac{1}{N} \sum_{i=1}^{N} -x_i^T y_i \tau^{-1} + \log (\max_{j\in B_i} \exp(x_i^T y_j {\tau}^{-1})) \\
&= \frac{1}{N} \sum_{i=1}^{N} {\tau}^{-1}(-x_i^T y_i \tau^{-1} + \max_{j\in B_i} x_i^T y_j).
\end{aligned}$
\label{lem:lb-global-2}
\end{lemma}

\subsubsection{Upper bounds on $\mathcal{L}^{Global} - \mathcal{L}^{Train}$}

We can now bound the gap between the global and training losses of the NT-Xent objective for a fixed batch set of batches $B$. The diagonal terms are included in both the global and training losses and will therefore not factor into characterizing the gap. 

\begin{theorem}[First upper bound on $\mathcal{L}^{Global} - \mathcal{L}^{Train}$]
An upper bound on $\mathcal{L}^{Global} - \mathcal{L}^{Train}$ for the NT-Xent objective using the lower bound from Lemma \ref{lem:lb-global-1} is: $\begin{aligned}
\mathcal{L}^{Global} - \mathcal{L}^{Train} \leq  \frac{1}{N} \sum_{i=1}^{N} &{\tau}^{-1} (\max_{j} x_i^T y_j- \min_{j \in B_i} x_i^T y_j) \\
&+ \log \frac{N}{k}.
\end{aligned}$
\label{thm:bound-gap-ntxent}
\end{theorem}

\begin{theorem}[Second upper bound on $\mathcal{L}^{Global} - \mathcal{L}^{Train}$]
An upper bound on $\mathcal{L}^{Global} - \mathcal{L}^{Train}$ for the NT-Xent objective using the lower bound from Lemma \ref{lem:lb-global-2} is: $\begin{aligned}
\mathcal{L}^{Global} - \mathcal{L}^{Train} \leq  \frac{1}{N} \sum_{i=1}^{N} &{\tau}^{-1} (\max_{j} x_i^T y_j - \max_{j \in B_i} x_i^T y_j) \\
&+ \log N.
\end{aligned}$
\label{thm:bound-gap-ntxent2}
\end{theorem}

\section{Minimizing $\mathcal{L}^{Global} - \mathcal{L}^{Train}$ as Quadratic Assignment over Batches}
\label{bound_qbap}

Note that from Theorems \ref{thm:bound-gap-ntxent} and \ref{thm:bound-gap-ntxent2} we have bounded the gap between $\mathcal{L}^{Global} - \mathcal{L}^{Train}$ without making any assumptions on data distribution or models. Additionally, we can see that the bounds are dependent on the batch assignments $j \in B_i$, batch size $k$, and total number of samples $N$. Note the losses we have characterized have been summed over samples in $X$ which we will denote as $\mathcal{L}^{Global}_X, \mathcal{L}^{Train}_X$ and consider losses summed over samples in $Y$ as $\mathcal{L}^{Global}_Y, \mathcal{L}^{Train}_Y$.

\subsection{Batches assignment as optimization over row permutations}

\noindent We will now rewrite our optimization problems over permutations $\pi \in \Pi_N$ instead of sets of batch assignments $\{B\}$, an equivalent formulation. First, recognize that our bounds are dependent only on batch assignments of negatives $j \in B_i$. Without loss of generality assume that batches are constructed sequentially after applying a row permutation $\pi \in \Pi_N$ on $X$ and $Y$. That is, batches are constructed over $\pi(X), \pi(Y)$ such that $j \in B_i \iff \lfloor \frac{j}{k} \rfloor = \lfloor \frac{i}{k} \rfloor$. Recognize that this batch construction can be written as a block diagonal matrix of the form $A \in \{0, 1\}^{N \times N}$ and $A_{i,j} = 1$ if $\lfloor \frac{j}{k} \rfloor = \lfloor \frac{i}{k} \rfloor$. Note this sequential constraint is not restrictive as it accommodates all possible batch assignments on $X, Y$ with the appropriate permutation $\pi \in \Pi_N$. When introducing a fixed sequential batching, we can rewrite the minimizer of the upper bound on $\mathcal{L}^{Global} - \mathcal{L}^{Train}$ from Theorems \ref{thm:bound-gap-ntxent} and \ref{thm:bound-gap-ntxent2} as an optimization problem over permutations $\pi \in \Pi_N$ on $X, Y$ rather than explicitly on $\{B\}$. The form of these optimizations problems are the Quadratic Bottleneck Assignment Problem and the Quadratic Assignment Problem \cite{koopmans}. These are well-known NP-Hard combinatorial optimization problems and in the following two sections we will discuss formulation and efficient approximations.


\subsection{Bounds related to Quadratic Bottleneck Assignment Problems}

\noindent The upper bound in Theorem \ref{thm:bound-gap-ntxent}, for the sum $(\mathcal{L}^{Global}_X - \mathcal{L}^{Train}_X) +(\mathcal{L}^{Global}_Y - \mathcal{L}^{Train}_Y)$, is minimized when the smallest inner product over in-batch negatives is maximized over $\pi \in \Pi_N$. Denote $Z_{ij} \triangleq \min \{x_i^T y_j, y_i^T x_j\}$ and $\odot$ as the Hadamard product. This is a QBAP, the proof of which is deferred to Appendix \ref{app-qbap-proof}, as we are interested in the minimizing the maximum value of the elementwise product of two symmetric matrices $\pi Z \pi^T$ and $A$.

\begin{theorem}[Formulation of QBAP for bound in Theorem \ref{thm:bound-gap-ntxent}]
\textit{The following Quadratic Bottleneck Assignment Problem, minimizes the upper bound provided in Theorem \ref{thm:bound-gap-ntxent} summed over $X$ and $Y$:} \[ \boxed{\min_{\pi \in \Pi_N} \max_{i, j} - A \odot \pi Z \pi^T.} \]
\label{thm:qap-bound-min-ip}
\end{theorem}

\subsection{Bounds related to Quadratic Assignment Problems}

The upper bound in Theorem \ref{thm:bound-gap-ntxent2}, for the sum $(\mathcal{L}^{Global}_X - \mathcal{L}^{Train}_X) +(\mathcal{L}^{Global}_Y - \mathcal{L}^{Train}_Y)$, is minimized when the sum of inner products over in-batch negatives is maximized over permutations $\pi \in \Pi_N$. This can be formulated equivalently as either the Frobenius inner product between the symmetric matrices $\pi (XY^T + YX^T) \pi^T$ and $A$ or the Trace of their product. These are QAPs, the proof of which is deferred to Appendix \ref{app-qap-proof}.

\begin{theorem}[Formulation of QAP for bound in Theorem \ref{thm:bound-gap-ntxent2}]
The following Quadratic Assignment Problem minimizes the upper bound in Theorem \ref{thm:bound-gap-ntxent2}:\[\boxed{\max_{\pi \in \Pi_N} Tr(A \pi (XY^T +  YX^T) \pi^T).} \]
\label{thm:qap-bound-max-ip}
\end{theorem}

Heuristics for both the QAP and QBAP in $\mathcal{O}(N^3)$ and $\tilde{\mathcal{O}}(N^2)$ time complexity respectively are well-known \cite{kuhn55, munkres57, edmonds72, Jonker88, cuthill-mckee}. In the next section, we will formulate approximate solutions to the QBAP in Theorem \ref{thm:qap-bound-min-ip} with $\mathcal{O}(Nk)$ space and $\tilde{\mathcal{O}}(N^2)$ time complexity.

\begin{figure*}
\centering
\includegraphics[width=\textwidth]{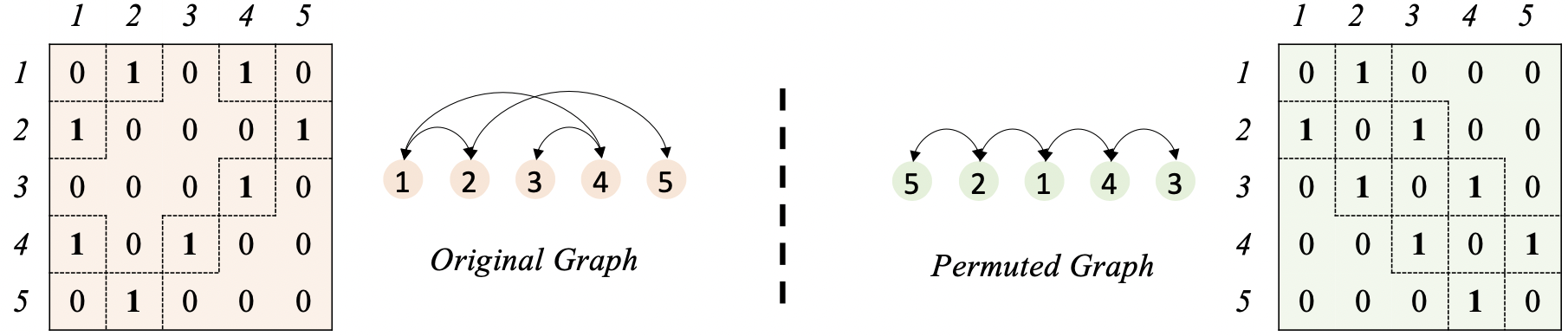}
\caption{Visual depiction of connection between minimizing graph and matrix bandwidths.}
\label{fig:graph_bandwidth}
\end{figure*}

\section{Global Contrastive Batch Sampling: Efficient approximations to the QBAP with Cuthill-McKee}
\label{sec_approx_qbap}
In practice, when $N$ is large it can be difficult to hold $XY^T$ in memory and approximation algorithms for the QAP problem \cite{kuhn55, munkres57, edmonds72, Jonker88} have $\mathcal{O}(N^3)$ complexity. Therefore, in optimizing over $\pi \in \Pi_N$ we will make two modifications in order to develop a $\mathcal{O}(Nk)$ space and $\tilde{\mathcal{O}}(N^2)$ worst-case time complexity approximation to the QBAP in Theorem \ref{thm:qap-bound-min-ip}. First, we will sparsify the matrix $XY^T$ on a quantile $q$ which censors values below the quantile to $0$. Secondly, we use an $\mathcal{\tilde{O}}(N^2)$ matrix bandwidth minimization heuristic commonly used for sparse matrix multiplication and decomposition \cite{cuthill-mckee} to efficiently attain an assignment over sample permutations.

\subsection{Approximating the QBAP: Sparsification and Matrix Bandwidth Minimization}

\noindent Previous literature \cite{burkard74} has shown the QBAP and Matrix Bandwidth Minimization Problem to be equivalent when the cost matrix is increasing in $|i-j|$ and \cite{burkard74} proposes thresholding in order to reduce coefficients in the matrix $XY^T$. First, we sparsify $XY^T$ on a threshold quantile $q$ as follows:
$$  (\tilde{XY}^T)_{i,j}
  =
  \left\{
  \begin{aligned}
    1, &\qquad& x_i^T y_j > q, i \neq j \\
    0,  &&  else.
  \end{aligned}
  \right.
$$

Note that there exists a minimal quantile $q^*$ which constructs a sparse matrix $\tilde{XY}^T$ that achieves the same solution as the dense matrix $XY^T$. This is due to the fact that since we are interested in maximizing the minimum inner product over in-batch negatives, the smallest values of $XY^T$ in each row are not of interest for the batch assignment objective.

\subsection{Approximating the QBAP: Cuthill-McKee Algorithm}

\begin{algorithm}[tb]
\caption{Cuthill-Mckee algorithm on sparse graph}
\label{alg:cap}
\begin{algorithmic}
\REQUIRE Sparse Adjacency Matrix $G \in \{0, 1\}^{N \times N}$
\STATE (1) Get peripheral vertex $v_i$ with lowest degree from the vertices in $G$. Set $\pi = [v_i]$. 
$\mathbf{Time \hspace{2mm} Compl.: \mathcal{O}(|E|) = \mathcal{O}((1-q)N^2) \approx \mathcal{O}(kN)}$
\STATE (2) Perform Breadth First Search on the Graph $G$ rooted at $v_i$ excluding elements in $\pi$. $\mathbf{Time \hspace{2mm} Compl.: \mathcal{O}(N)}$
\STATE (3) Label each vertex, other than $v_i$, on their distance from $v_i$, creating "levels". $\mathbf{Time \hspace{2mm} Compl.:\mathcal{O}(N)}$
\STATE (4) Order vertices by level, tiebreaker of ascending vertex degree and append the first item to $\pi$. $\mathbf{Time \hspace{2mm} Compl.: \mathcal{O}(N log(N))}$
\STATE (5) If $|\pi| < N$, return to Step (2) and repeat this process with the most recently added vertex as the root. 
\STATE (6) Return permutation $\pi = [i_0, i_1, \dots, i_n]$
\end{algorithmic}
\end{algorithm}

On this sparsified matrix $\tilde{XY}^T$, we should seek to maximize the number of nonzero values in $ \pi (\tilde{XY}^T + (\tilde{XY}^T)^T )\pi^T \odot A$ to minimize the upper bound in \ref{thm:qap-bound-min-ip}. The QBAP formulations and the Matrix Bandwidth Minimization problem are approximately equivalent due to the fixed sequential batching which assigns batches along the main diagonal of $\pi XY^T \pi^T$. Since $\{B\}$ is comprised of blocks on the main diagonal, minimizing the dispersion of non-zero entries, after sparsification, from the main diagonal will maximize the probability of large inner product values within batches. As a result, our algorithm for minimizing $(\mathcal{L}^{Global}_X - \mathcal{L}^{Train}_X) +(\mathcal{L}^{Global}_Y - \mathcal{L}^{Train}_Y)$, GCBS, is an $\mathcal{\tilde{O}}(N^2)$ relaxation to the bound in Theorem \ref{thm:qap-bound-min-ip}. In Algorithm \ref{alg:cap} we detail the Cuthill-Mckee algorithm for matrix bandwidth minimization \cite{cuthill-mckee} along with worst case runtimes for each step. 

Additionally, we note that the Cuthill-Mckee algorithm has been extensively applied in graph bandwidth problems. In the directed unweighted graph setting, a linear graph arrangement \cite{graph-bandwidth} is applied on an adjacency matrix $G \in \{0, 1\}^{N \times N}$ such that each node is placed at the corresponding row integer value $i$ on the x-axis with edges to nodes $j$ if $G_{ij} \neq 0$. The objective in this case is to minimize the length of the longest edge. By viewing the batch assignment problem in this graphical setting, one can recognize that in our sparse implementation GCBS seeks to minimize the distance between nodes $i, j$ where $x_i^Ty_j$ is a relatively large inner product. This connection between minimizing graph and matrix bandwidths is shown visually in Figure \ref{fig:graph_bandwidth}.

\begin{table*}[!ht]
\centering
\begin{tabular}{lc|c|c c c c c c c} \toprule
    {Model} & CosQA & AdvTest & Ruby & JS & Go & Python & Java & PHP & CSN Avg  \\ \midrule
    RoBERTa             & 60.3 & 18.3 & 58.7 & 51.7 & 85.0 & 58.7 & 59.9 & 56.0 & 61.7  \\
    CodeBERT            & 65.7 & 27.2 & 67.9 & 62.0 & 88.2 & 67.2 & 67.6 & 62.8 & 69.3  \\
    GraphCodeBERT       & 68.4 & 35.2 & 70.3 & 64.4 & 89.7 & 69.2 & 69.1 & 64.9 & 71.3 \\
    SYNCoBERT           & -    & 38.3 & 72.2 & 67.7 & 91.3 & 72.4 & 72.3 & 67.8 & 74.0  \\ 
    PLBART              & 65.0 & 34.7 & 67.5 & 61.6 & 88.7 & 66.3 & 66.3 & 61.1 & 68.5  \\
    CodeT5-base         & 67.8 & 39.3 & 71.9 & 65.5 & 88.8 & 69.8 & 68.6 & 64.5 & 71.5 \\
    \midrule
    UniXcoder           & 70.1 & 41.3 & 74.0 & 68.4 & 91.5 & 72.0 & 72.6 & 67.6 & 74.4 \\
    \rowcolor{Gainsboro!60}
    \hspace{5mm} - with GCBS    & \textbf{71.1} (+1.0) & \textbf{43.3} (+2.0) & \textbf{76.7} & \textbf{70.6} & \textbf{92.4} & \textbf{74.6} & \textbf{75.3} & \textbf{70.2} & \textbf{76.6} (\textbf{+2.2}) \\
    \bottomrule
\end{tabular}
\caption{The performance comparison of supervised models along with a comparison of the best performing model (UniXcoder) \cite{guo-etal-2022-unixcoder} when using GCBS vs the standard Random Sampling. The reported score is Mean Reciprical Rank magnified by a factor of 100. GCBS improves previous best MRR when used with UniXcoder by 2.2 points achieving new state-of-the-art results (Row shaded gray).}
\label{table:code-search}
\end{table*}

\begin{table*}[!ht]
\centering
\begin{tabular}{lc|c|c|c|c|c|c|c} \toprule
    {Model} & STS12 & STS13  & STS14 & STS15 & STS16 & STS-B & SICK-R & Avg  \\ \midrule
    SBERT$_{base}$ & 70.97 & 76.53 & 73.19 & 79.09 & 74.30 & 77.03 & 72.91 & 74.89  \\
    SBERT$_{base}$-flow  & 69.78 & 77.27 & 74.35 & 82.01 & 77.46 & 79.12 & 76.21 & 76.60  \\
    SBERT$_{base}$-whitening & 69.65 & 77.57 & 74.66 & 82.27 & 78.39 & 79.52 & 76.91 & 77.00 \\
    ConSERT-BERT$_{base}$  & 74.07 & 83.93 & 77.05 & 83.66 & 78.76 & 81.36 & 76.77 & 79.37  \\ 
    \midrule
    SimCSE-BERT$_{base}$ & 75.30 & 84.67 & 80.19 & 85.40 & 80.82 & 84.25 & 80.39 & 81.57   \\
    \rowcolor{Gainsboro!60}
    \hspace{5mm} - with GCBS & 75.81 & 85.30 & 81.12 & 86.58 & 81.68 & 84.80 & 80.04 & 82.19 (+0.62)  \\
    PromCSE-BERT$_{base}$ & 75.96 & 84.99 & 80.44 & 86.83 & 81.30 & 84.40 & 80.96 & 82.13 \\
    \rowcolor{Gainsboro!60}
    \hspace{5mm} - with GCBS & 75.20 & 85.00 & 81.00 & 86.82 & 82.55 &    84.76     &      79.95      & 82.18 (+0.05) \\
    SimCSE-RoBERTa$_{base}$ & 76.53 & 85.21 & 80.95 & 86.03 & 82.57 & 85.83 & 80.50 & 82.52  \\
    \rowcolor{Gainsboro!60}
    \hspace{5mm} - with GCBS & 76.94 & 85.64 & 81.87 & 86.84 & 82.78 & 85.87 & 80.68 & 82.95 (+0.43) \\
    PromCSE-RoBERTa$_{base}$ & 77.51 & 86.15 & 81.59 & 86.92 & 83.81 & 86.35 & 80.49 & 83.26 \\
    \rowcolor{Gainsboro!60}
    \hspace{5mm} - with GCBS & 77.33 & 86.77 & 82.19 & 87.57 & 84.09 &    86.78     &      80.05      & 83.54 (+0.28) \\
    SimCSE-RoBERTa$_{large}$ & 77.46 & 87.27 & 82.36 & 86.66 & 83.93 & 86.70 & 81.95 & 83.76   \\
    \rowcolor{Gainsboro!60}
    \hspace{5mm} - with GCBS & 78.90 & 88.39 & 84.18 & 88.32 & 84.85 & 87.65 & 81.27 & 84.79 (+1.03)  \\
    PromCSE-RoBERTa$_{large}$ & 79.56 & 88.97 & 83.81 & 88.08 & 84.96 & \textbf{87.87} & \textbf{82.43} & 85.10 \\
    \rowcolor{Gainsboro!60}
    \hspace{5mm} - with GCBS & \textbf{80.49} & \textbf{89.17} & \textbf{84.57} & \textbf{88.61} & \textbf{85.38} &    \textbf{87.87}     &      81.49      & \textbf{85.37} (\textbf{+0.27})  \\
    \bottomrule
\end{tabular}
\caption{The performance comparison of supervised models along with a comparison of the best performing models, SimCSE \cite{gao2021simcse} and PromCSE \cite{jiang2022promcse}, with and without GCBS. The reported score is Spearman correlation magnified by a factor of 100. For RoBERTa$_{large}$ backbone models, GCBS improves previous best Spearman correlation when used with SimCSE by 1.03 points and PromCSE by 0.27 points achieving new state-of-the-art results.  }
\label{table:sent-embed}
\end{table*}


\section{Experimentation}
\label{Exps}

In this section, we detail experiments for sentence embedding and code-search tasks when using GCBS instead of the standard Random Sampling. We find that GCBS improves state-of-the-art performance for both tasks while requiring minimal code changes. Specification of hyperparameters are included in Appendix Section \ref{Hparams}. 

\subsection{Code Search Experiments}

Semantic code search is an important problem in representation learning that jointly embeds programming and natural languages \cite{husain2019codesearchnet}. In this task, one is concerned with returning relevant code when given a natural language query. This is a problem of great interest due to the potential for aiding programmers when developing code and possesses challenges in aligning highly technical and abbreviated language with the programming language modality. Recently, models for this task have been improved using contrastive learning \cite{guo-etal-2022-unixcoder} by enforcing sequence embeddings for code and their corresponding natural language comments to have large inner products relative to unrelated natural language comments. GCBS provides further gains and achieves state-of-the-art performance when used with well-performing contrastive learning models, UniXcoder \cite{guo-etal-2022-unixcoder} as shown in Table \ref{table:code-search}.

\subsection{Sentence Embedding Experiments}

Recently, contrastive learning approaches, which enforce that pretrained sentence embeddings for mined pairs of similar sentences have large inner products relative to the inner products of random pairs of sentences, have provided state of the art performance for sentence embedding tasks. GCBS provides further gains and achieves state-of-the-art performance when used with well-performing contrastive learning models, SimCSE \cite{gao2021simcse} and PromCSE \cite{jiang2022promcse}, as shown in Table \ref{table:sent-embed}

\subsection{Self-Supervised Image Classification Experiments}

Additionally, we conduct experiments on vision datasets with strong and extensive self-supervised image classification baselines across a variety of methods from past literature. In particular, we evaluate on the Imagenette and Cifar10 vision classification datasets obtained from Lightly AI \footnote{\url{https://docs.lightly.ai/self-supervised-learning/getting_started/benchmarks.html}}. 

On these tasks, we find GCBS to be beneficial when implemented both with Moco (e.g., leveraging a memory bank) \cite{moco_v1} and SimCLR (e.g., using various (learnable) data augmentations in training) \cite{simclr}. Performance on the Imagenette and Cifar10 experiments are provided below in Table \ref{table:imagenette-gcbs} and Table \ref{table:cifar10-gcbs} respectively. 

\begin{table}[h!]
\centering
\begin{tabular}{lc|c} \toprule
    {Model} & Dataset & Test Accuracy (kNN) \\ \midrule
    SimCLR & Imagenette & 89.2 \\
    \rowcolor{Gainsboro!60}
    SimCLR w/ GCBS & Imagenette & \textbf{90.9 (+1.7)} \\
    Moco & Imagenette & 87.6 \\
    \rowcolor{Gainsboro!60}
    Moco w/ GCBS & Imagenette & 88.9 (+1.3) \\
    \bottomrule
\end{tabular}
\caption{The performance of Moco and SimCLR models \cite{moco_v1, simclr} with and without GCBS on self-supervised image classification for the Imagenette dataset. The reported score is the Test Accuracy using kNN.}
\label{table:imagenette-gcbs}
\end{table}

\begin{table}[h]
\centering
\begin{tabular}{lc|c} \toprule
    {Model} & Dataset & Test Accuracy (kNN) \\ \midrule
    SimCLR & Cifar10 & 87.5 \\
    \rowcolor{Gainsboro!60}
    SimCLR w/ GCBS & Cifar10 & 89.8 (+2.3) \\
    Moco & Cifar10 & 90.0 \\
    \rowcolor{Gainsboro!60}
    Moco w/ GCBS & Cifar10 & \textbf{90.3 (+0.3)} \\
    \bottomrule
\end{tabular}
\caption{The performance of Moco and SimCLR models \cite{moco_v1, simclr} with and without GCBS on self-supervised image classification for the Imagenette dataset. The reported score is the Test Accuracy using kNN.}
\label{table:cifar10-gcbs}
\end{table}

\section{Discussion}
\label{Discussion}

In this section, we analyze the loss of contrastive learning models when using GCBS compared to Random Sampling. We find that Global Contrastive Batch Sampling empirically reduces the gap $\mathcal{L}^{Global}- \mathcal{L}^{Train}$, as intended, by $40\%$ in Code Search Net (Ruby) experiments. Additionally, runtime for each epoch when using GCBS is approximately $60\%$ that of the most minimal Hard Negative Mining implementation.  

\subsection{Runtime comparison}
\label{sec:runtimes}

\begin{table}[!ht]
\small
\centering
\begin{tabular}{lc|c|c} \toprule
     & \multicolumn{3}{c}{Code Search} \\
    \hline
    Step & Random & GCBS & Hard Negative (1) \\ \midrule
    Fwd+Bkwd Pass & 381.45 & 381.45 & 762.9 (2x batches) \\
    Add'l Fwd Pass & - & 118.51 & 118.51 \\
    \hline
    Comp. k-NN & - & - & 1.19  \\
    GCBS & - & 2.31 & -  \\
    \hline
    Total Time (s) & 381.45 & 502.27 & 882.61  \\
    \bottomrule
\end{tabular}
\caption{Runtime in seconds per epoch for Random Sampling, GCBS, and Hard Negative (1) for the Code Search Net (Ruby) dataset $N=24,927, k=64$ with the UniXcoder model.}
\label{runtimes-cs}
\end{table}

We provide runtime comparisons for GCBS, Random Sampling and Hard Negative (1), which mines one hard negative per sample and recomputation of nearest neighbors once an epoch. We find that GCBS is more efficient per epoch than this minimal implementation of Hard Negative mining. Runtimes were calculated using a single NVIDIA A100 GPU with CUDA 11.6 and PyTorch version 1.11.0, 52GB RAM, and 4 vCPUs. Runtime statistics for the Code Search Net (Ruby) dataset with the UniXcoder model in Table \ref{runtimes-cs} and the SNLI+MNLI (entailment+hard neg) dataset for sentence embedding with the BERT$_{base}$ model are shown in Table \ref{runtimes-se}.

\begin{table}[ht!]
\small
\centering
\begin{tabular}{lc|c|c} \toprule
    & \multicolumn{3}{c}{Sentence Embedding} \\
    \hline
    Step & Random & GCBS & Hard Negative (1) \\ \midrule
    Fwd+Bkwd Pass & 442.26 & 442.26 & 884.52 (2x batches) \\
    Add'l Fwd Pass & - & 370.31 & 370.31 \\
    \hline
    Comp. k-NN  & - & - & 225.99 \\
    GCBS & - & 140.32 & - \\
    \hline
    Total Time (s)  & 442.26 & 965.03 & 1480.82 \\
    \bottomrule
\end{tabular}
\caption{Runtime in seconds per epoch for Random Sampling, GCBS, and Hard Negative (1) for the SNLI+MNLI (entailment+hard neg) dataset $N=275,602, k=256$ for sentence embedding with the Bert-base-uncased model.}
\label{runtimes-se}
\end{table}

\subsection{Global, Training losses for GCBS vs Random Sampling}

\begin{figure}[!ht]
\centering
\includegraphics[width=0.45\textwidth]{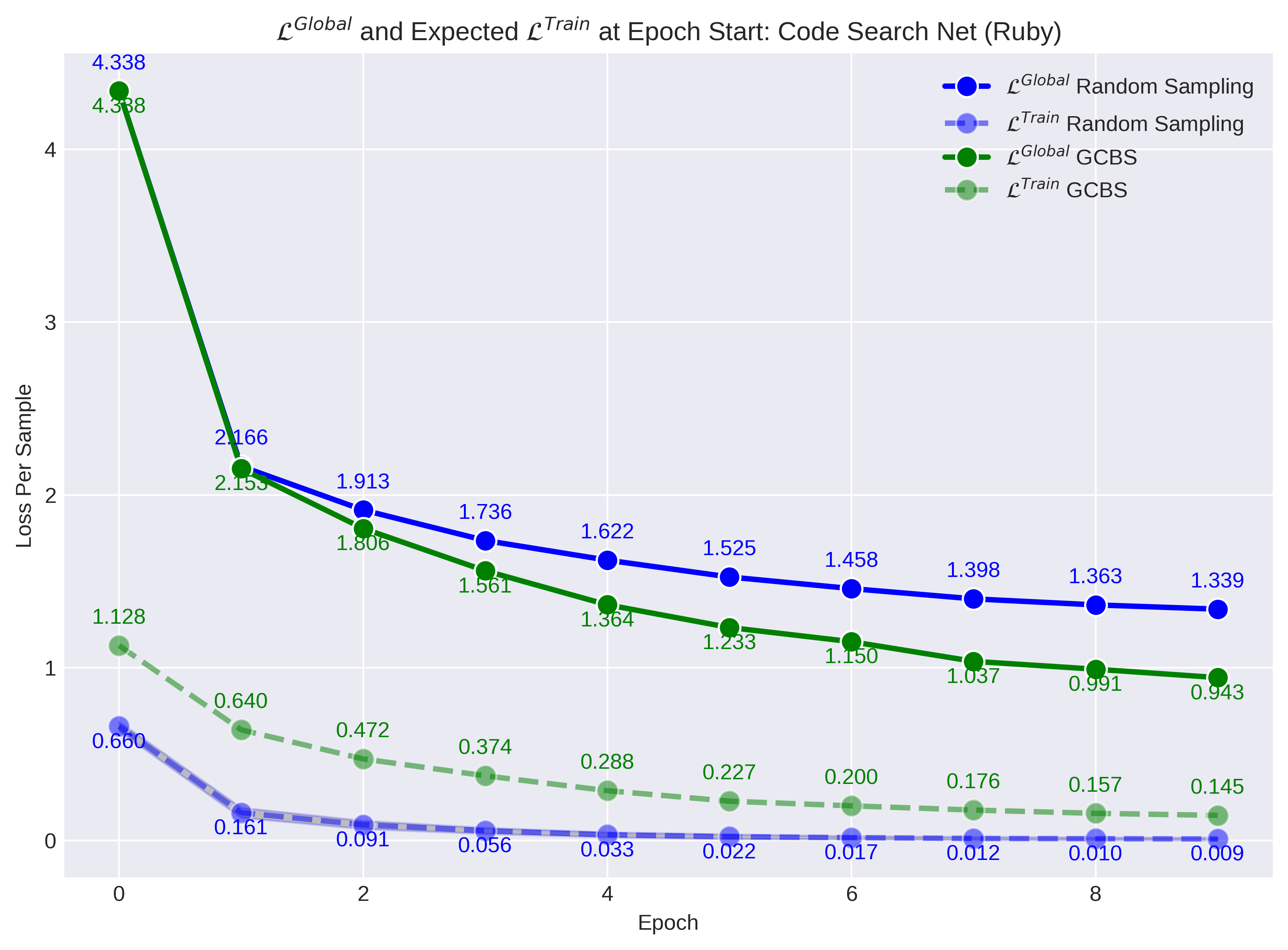}
\caption{$\mathcal{L}^{Global}$ and Expected $\mathcal{L}^{Train}$ at the start of each epoch for Random Sampling and GCBS on the Code Search Net (Ruby) dataset with the UniXcoder model.}
\label{gap_losses}
\end{figure}

Empirically, we verify our theoretical contributions that Matrix Bandwidth Minimization will reduce the gap between $\mathcal{L}^{Global} - \mathcal{L}^{Train}$. To perform this study, we calculate the loss for in-batch negatives and the loss over all negatives for each sample when using either Random Sampling and GCBS for the Code Search Net (Ruby) dataset with the UniXcoder model. As shown in Figure \ref{gap_losses}, we find that using GCBS reduces the gap $\mathcal{L}^{Global}- \mathcal{L}^{Train}$ by ~40\% when compared to Random Sampling on the final epoch. Additionally, the total loss over all samples is reduced by 30\% and, as shown in Appendix Section \ref{val_perf}, yields stronger validation/test performance.

\section{Conclusion}
\label{Conclusion}
In this paper, we introduced Global Contrastive Batch Sampling (GCBS), an efficient algorithm for better approximating global losses in contrastive learning through global batch assignments. GCBS is an approximation for quadratic assignment problems we prove characterize upper bounds for the gap between global and training losses in contrastive learning. Unlike previous approaches using hard negative mining, GCBS does not increase the training length of epochs by oversampling and is more efficient compared the most minimal hard negative mining approaches. We evaluate GCBS on sentence embedding and code search tasks and achieve state-of-the-art performance in both settings. Our method provides an efficient alternative to hard negative mining that is simple to implement, does not maintain additional data structures during training, provides strong performance, and performs global batch assignments.

\section*{Acknowledgements}

The authors would like to thank Shi Dong and Junheng Hao for their helpful comments and discussions. VS would like to thank Ryan Theisen for helpful discussions on early versions of this paper. 

\bibliography{example_paper}
\bibliographystyle{icml2023}

\newpage
\appendix
\onecolumn

\section{Derivation of Proofs for Theorems \ref{thm:qap-bound-min-ip} and \ref{thm:qap-bound-max-ip}}
\label{proofs}

In this section, we provide proof derivations of Theorems \ref{thm:qap-bound-min-ip} and \ref{thm:qap-bound-max-ip}.

\subsection{Proof of Theorem \ref{thm:qap-bound-min-ip}}
\label{app-qbap-proof}


We show that the formulation of the gap between the Global and Training contrastive losses $\mathcal{L}^{Global} - \mathcal{L}^{Train}$ when using the translation identity lower bound for Log-Sum-Exp \cite{Nielsen-Sun} is approximated as a Quadratic Bottleneck Assignment Problem (QBAP). This optimization problem is associated with the lower bound in Theorem \ref{thm:bound-gap-ntxent}.

Since this formulation is not equivalent over $X$ and $Y$, we will first denote $\mathcal{L}^{Global} - \mathcal{L}^{Train}_X$ and $\mathcal{L}^{Global} - \mathcal{L}^{Train}_Y$ as the respective gaps over $X$ and $Y$ when using the translation identity lower bound on $\mathcal{L}^{Train}_X, \mathcal{L}^{Train}_Y$:

$$\begin{aligned}
\mathcal{L}^{Global}_X - \mathcal{L}^{Train}_X \leq& \frac{1}{N} \sum_{i=1}^{N} {\tau}^{-1} (\max_{j} x_i^T y_j -  \min_{j \in B_i} x_i^T y_j) + \log \frac{N}{k} & \\
\mathcal{L}^{Global}_Y - \mathcal{L}^{Train}_Y \leq& \frac{1}{N} \sum_{i=1}^{N} {\tau}^{-1} (\max_{j} y_i^T x_j -  \min_{j \in B_i} y_i^T x_j) + \log \frac{N}{k} & \\
\end{aligned}$$

Then we will minimize the optimization problem $\min_B \mathcal{L}^{Global}_X + \mathcal{L}^{Global}_Y - \mathcal{L}^{Train}_X - \mathcal{L}^{Train}_Y$ in order to equally weigh the selection of informative samples for both $X$ and $Y$. Lastly, denote $Z_{ij} \triangleq \min \{x_i^T y_j, y_i^T x_j\}$ and $\odot$ as the Hadamard product.

\begin{proof}
To formulate $\min_B \mathcal{L}^{Global}_X + \mathcal{L}^{Global}_Y - \mathcal{L}^{Train}_X - \mathcal{L}^{Train}_Y$ with the translation identity lower bound for Log-Sum-Exp as a QBAP, we need only use the fact that $\sum_i^N x_i \geq N \min_i x_i$.

\begin{equation}
\begin{aligned}
\arg \min_B \mathcal{L}^{Global}_X + \mathcal{L}^{Global}_Y - \mathcal{L}^{Train}_X - \mathcal{L}^{Train}_Y =& \arg \max_{B} \sum_{i=1}^{N} \min_{j \in B_i} x_i^T y_j  + \min_{j \in B_i} y_i^T x_j & \\
\geq& 2 \arg \max_{B} \sum_{i=1}^{N}  \min\{ \min_{j \in B_i} x_i^T y_j, \min_{j \in B_i} y_i^T x_j\} & \\
=& 2 \arg \max_{B} \sum_{i=1}^{N} \min_{j \in B_i} Z_{ij} & \\
\geq& 2N \arg  \max_{B} \min_{i, j \in B} Z_{ij}  &\\
=& 2N \arg \min_{\pi \in \Pi_N} \max_{i, j} -A \odot \pi Z \pi^T.  & \\
\end{aligned}
\end{equation}
\end{proof}

\subsection{Proof of Theorem \ref{thm:qap-bound-max-ip}}
\label{app-qap-proof}

We show that the formulation of the gap between the Global and Training contrastive losses $\mathcal{L}^{Global} - \mathcal{L}^{Train}$ when using the standard lower bound for Log-Sum-Exp \cite{OptModels} is approximated as a Quadratic Assignment Problem (QAP). This optimization problem is associated with the lower bound Theorem \ref{thm:bound-gap-ntxent2}.

Since this formulation is not equivalent over $X$ and $Y$, we will first denote $\mathcal{L}^{Global} - \mathcal{L}^{Train}_X$ and $\mathcal{L}^{Global} - \mathcal{L}^{Train}_Y$ as the respective gaps over $X$ and $Y$ when using the standard lower bound on $\mathcal{L}^{Train}_X, \mathcal{L}^{Train}_Y$:

$$\begin{aligned}
\mathcal{L}^{Global}_X - \mathcal{L}^{Train}_X \leq & \frac{1}{N} \sum_{i=1}^{N} {\tau}^{-1} (\max_{j} x_i^T y_j -  \max_{j \in B_i} x_i^T y_j) + \log N \\
\mathcal{L}^{Global}_Y - \mathcal{L}^{Train}_Y \leq & \frac{1}{N} \sum_{i=1}^{N} {\tau}^{-1} (\max_{j} y_i^T x_j -  \max_{j \in B_i} y_i^T x_j) + \log N \\
\end{aligned}$$

Then we will minimize the optimization problem $\min_B \mathcal{L}^{Global}_X + \mathcal{L}^{Global}_Y - \mathcal{L}^{Train}_X - \mathcal{L}^{Train}_Y$ in order to equally weigh the selection of informative samples for both $X$ and $Y$. Lastly, denote $\odot$ as the Hadamard product.

\begin{proof}
To formulate $\min_B \mathcal{L}^{Global}_X + \mathcal{L}^{Global}_Y - \mathcal{L}^{Train}_X - \mathcal{L}^{Train}_Y$ as a QAP, we need only use the fact that $\max_{\{i \in 1, 2, \dots, k\}} x_i \geq \frac{1}{k} \sum_i^k x_i$.
\begin{equation}
\begin{aligned}
\arg \min_B \mathcal{L}^{Global}_X + \mathcal{L}^{Global}_Y - \mathcal{L}^{Train}_X - \mathcal{L}^{Train}_Y =& \arg \max_{B} \sum_{i=1}^{N} \max_{j \in B_i} x_i^T y_j + \max_{j \in B_i} y_i^T x_j \\
\geq& \frac{1}{k} \arg \max_{B} \sum_{i=1}^{N} \sum_{j \in B_i} x_i^T y_j + \frac{1}{k} \max_{B} \sum_{i=1}^{N} \sum_{j \in B_i} y_i^T x_j \\
=& \frac{1}{k} \arg \max_{\pi \in \Pi_N} Tr(A \pi XY^T \pi^T) + Tr((A \pi XY^T \pi^T)^T ) \\
=& \frac{1}{k} \arg \max_{\pi \in \Pi_N} Tr(A \pi (XY^T +  YX^T) \pi^T ).
\end{aligned}
\end{equation}
\end{proof}

\section{Expected loss values at Epoch Start for Random Sampling (10000 trials) in Code Search Net (Ruby)}
\begin{figure}[!ht]
\centering
\includegraphics[width=0.79\textwidth]{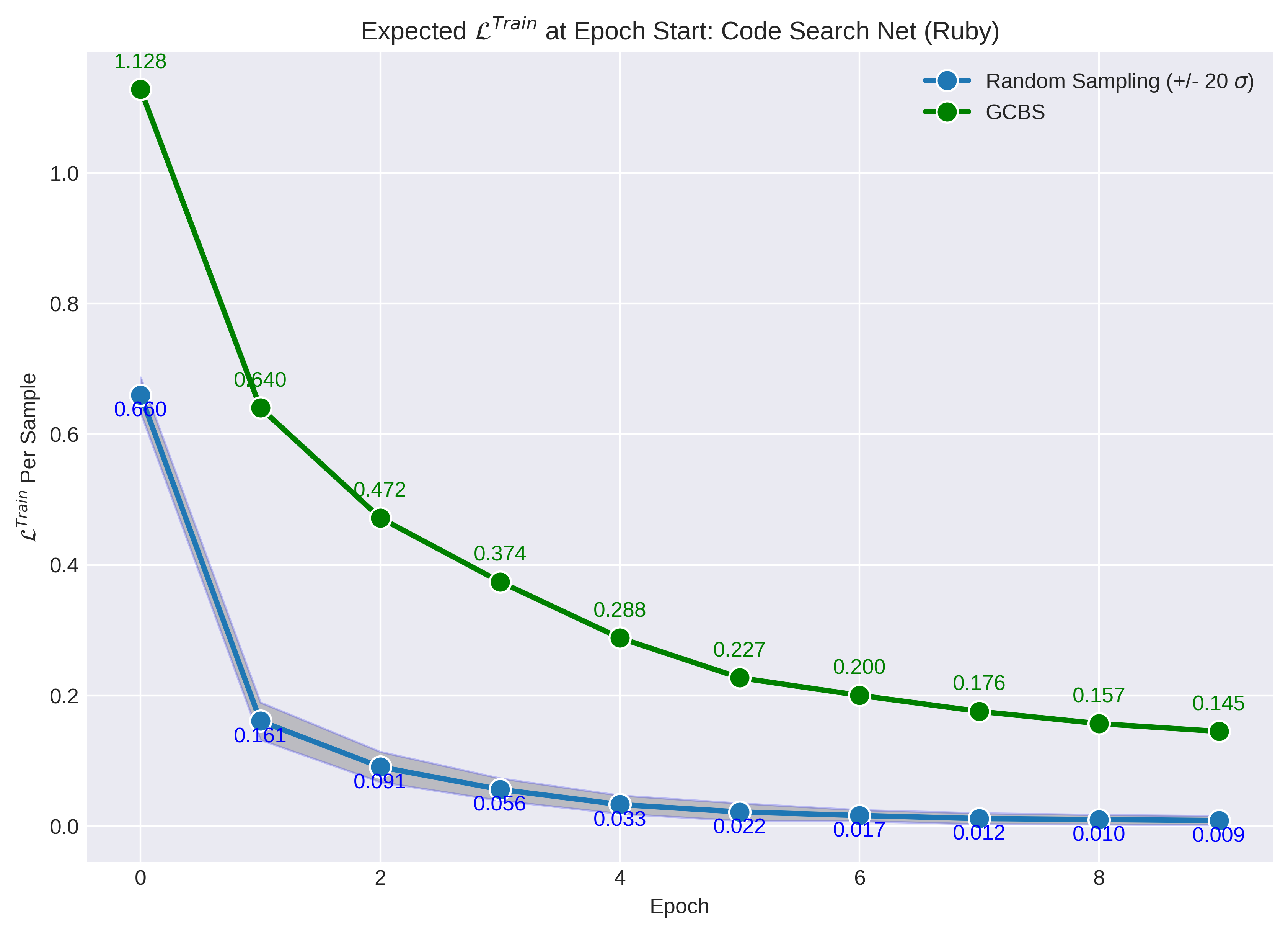}
\caption{Expected $\mathcal{L}^{Train}$ for Random Sampling ($\pm 20\sigma$) and GCBS on the Code Search Net (Ruby) dataset with the UniXcoder model.}
\label{std_dev_random}
\end{figure}

We calculate the expected loss for Random Sampling over 10,000 random batch assignments and compare these loss values to GCBS. The expected loss values for Random Sampling is clearly differentiated from Global Contrastive Batch Sampling even when compared with the mean over the 10,000 assignments plus $20$ standard deviations as shown in Figure \ref{std_dev_random}. As a result, the loss incurred by GCBS is a better proxy of the global loss even compared to the largest loss incurred among 10,000 random assignments. Empirically, this shows that GCBS provides improvements in Global batch assignment that are unlikely to be obtained by selecting across random assignments. 

\section{Validation Performance comparison for GCBS, Random Sampling, and Hard Negative Mining}
\label{val_perf}
In this section, we provide validation and test performance for GCBS, Random Sampling, and Hard Negative Mining for the Code Search Net (Ruby) dataset with the UniXcoder model. We find that GCBS provides validation and test performance improvements compared to both Random Sampling and Hard Negative (1). In particular, the gap between test performance for GCBS and Hard Negative (1) is greater than that of Hard Negative (1) and Random Sampling.

\begin{figure}[!ht]
\centering
\includegraphics[width=0.79\textwidth]{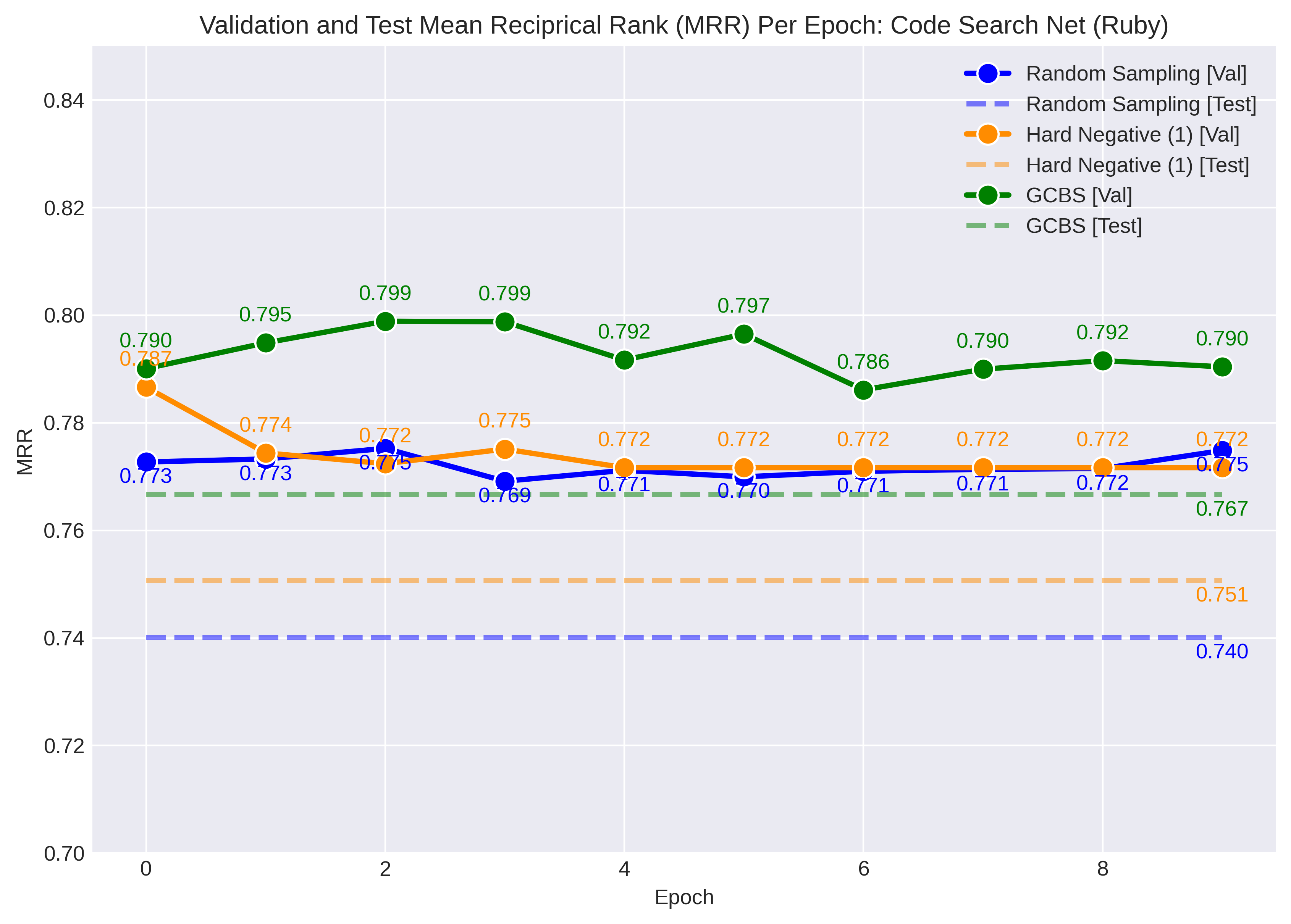}
\caption{Validation and Test Performance for Random Sampling, Hard Negative (1), and GCBS on the Code Search Net (Ruby) dataset with the UniXcoder model \cite{guo-etal-2022-unixcoder}.}
\label{val_test_perf}
\end{figure}

\section{Expected positive class Softmax probability at epoch start for GCBS, Random Sampling, and Global}
\label{softmax_prob}

In this section, we provide the expected softmax probability at the start of each epoch across in-batch negatives when using GCBS and Random Sampling and across all negative samples (i.e. Global setting) for the Code Search Net (Ruby) task with the UniXcoder model. We find that GCBS better approximates the softmax probability of positive classes compared to random sampling and that random sampling results in loss saturation within a small number of epochs.

\begin{figure}[!ht]
\centering
\includegraphics[width=0.79\textwidth]{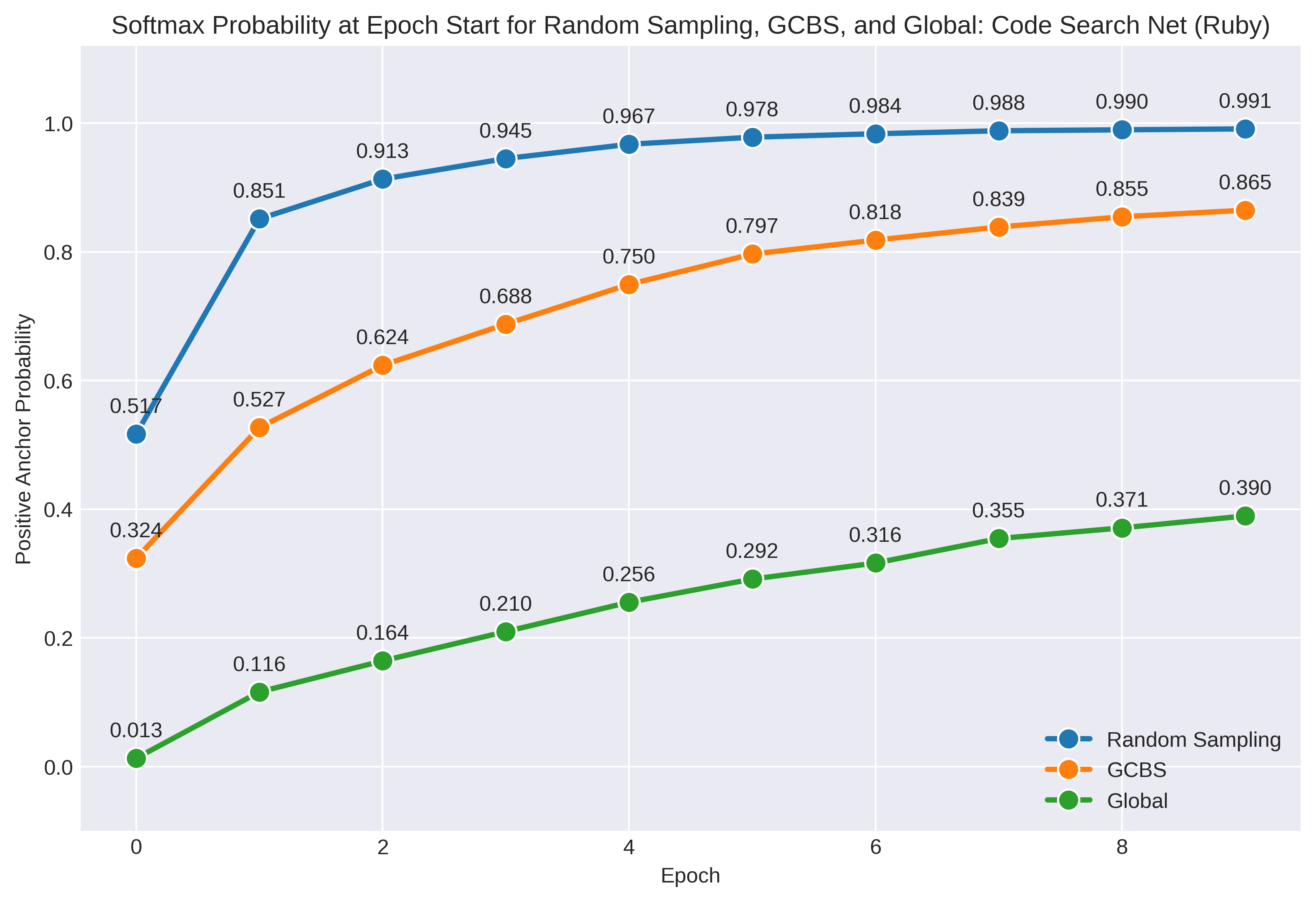}
\caption{Expected positive anchor Softmax Probability for Code Search Net (Ruby) task using the UniXcoder model \cite{guo-etal-2022-unixcoder}.}
\label{softmax_prob_fig}
\end{figure}

\section{Complexity Analysis of Global Contrastive Batch Sampling}
\label{complexity}

\subsection{Quantile Estimation}
First, it is necessary to compute the value at the quantile $q$ in order to sparsify $XY^T$. For large datasets in our experiments, this operation is estimated over chunks of $XY^T$ and the median quantile value over the chunks is used. For each chunk of size $l$, this requires computing values of $XY^T$, performing a sort on these values and getting the index of the sorted values for the specified quantile. We denote matrix multiplication time complexity between two matrices as $MM(\cdot, \cdot)$ and show that this estimation has time complexity approximately equivalent to the matrix multiplication $XY^T$. We make the assumption that $d \geq \log(Nl)$. 

\begin{equation}
\begin{aligned}
\text{Space Complexity} =& \underbrace{Nl^{-1}}_{\text{Chunk quantile values}} + \underbrace{Nl}_{\text{items in each chunk}} = \mathcal{O}(Nl) \\ 
\text{Time Complexity} =& 2Nl^{-1}(\underbrace{\mathcal{O}(Nl \log(Nl))}_{\substack{\text{Sort inner products}}} + \underbrace{MM(Nd, ld)}_\text{Calculate inner products}) = \mathcal{O}(N^2d)
\end{aligned}
\end{equation}

\subsection{Optimizing over row permutations of X, Y}
After estimating a value at which to sparsify, we need to get a sparsified similarity matrix $\tilde{XY}^T$, construct a sparse adjacency matrix and run the Cuthill McKee algorithm. We assume that $N (1-q)$, or the expected number of entries in each row is a small multiple of the batch size $k$. First, we detail the space and time complexity for constructing $\tilde{XY}^T$:

\begin{equation}
\begin{aligned}
\text{Space Complexity} =&  \underbrace{3N^2(1-q)}_{\text{Row, Column, and Data values of $\tilde{XY}^T$}} = \mathcal{O}(Nk) \\ 
\text{Time Complexity} =& \underbrace{Nl^{-1} MM(Nd, ld)}_\text{Calculate inner products and threshold} = \mathcal{O}(N^2d)
\end{aligned}
\end{equation}

The space and time complexity for running the Cuthill McKee algorithm on $\tilde{XY}^T$ is detailed below, we assume the implementation from \cite{linCM} is used which provides runtime bounded by $N^2$ up to logarithmic factors. 

\begin{equation}
\begin{aligned}
\text{Space Complexity} =& \underbrace{3N^2(1-q)}_{\text{Row, Column, and Data values of $\tilde{XY}^T$}} = \mathcal{O}(Nk)  \\ 
\text{Time Complexity} =& \underbrace{\mathcal{O}(mN \log(m)) }_{\text{Cuthill-Mckee Runtime}} \leq \underbrace{\mathcal{O}(N^2 \log(N)) }_{\text{Worst case, $m=N$}}
\end{aligned}
\end{equation}

Note that $m$ is the maximum degree over nodes and for a large quantile value will typically be smaller than $N$. We find that Global Contrastive Batch Sampling incurs $\mathcal{O}(Nk)$ space complexity and $\mathcal{O}(N^2 d)$ time complexity.

\section{Implementation in PyTorch}
\label{GCBS-code}
In this section, we detail efficient implementation of GCBS in PyTorch. Our implementation computes a permutation over samples $\pi$ at the beginning of each epoch, requires less than 50 lines of code, makes no changes to the model being trained, and does not maintain external data structures after being run between epochs.

The PyTorch pseudocode for the implementation of GCBS is contained below. In the case where $XY^T$ cannot be held in memory, the value of the quantile $q$ can be approximated over subsamples of entries from $XY^T$ and the sparse matrix $\tilde{XY}^T$ can be constructed similarly.
\begin{lstlisting}
def compute_perm_bandwidth_min(X, Y, quantile_thresh = 0.999):
      # (1) Normalize representations.
      X, Y = normalize(X), normalize(Y)
      
      # (2) Get value at quantile threshold on the inner product matrix.
      quantile_thresh = torch.quantile(X @ Y.T, quantile_thresh)
      
      # (3) Get inner product matrix hard thresholded on quantile.
      row, col, data = [], [], []
      
      # Get rows and columns of indices > estimated quantile value
      ret = ((X @ Y.T).flatten() > quantile_thresh).nonzero
      row += ((ret - (ret % num_samples))/num_samples).tolist()
      col += (ret % num_samples).tolist()
      data += [1.0 for _ in range(len(ret))]

      # (4) Get perm which minimizes bandwidth of sparsified matrix with Cuthill-McKee.
      permutation = list(cuthill_mckee(sparse_matrix((data, (row, col)), 
                                       shape=(num_samples, num_samples))))
      return permutation

\end{lstlisting}

In the next code block, we provide PyTorch pseudocode which, when inserted at the beginning of each epoch, will call the previous method and apply the permutation over samples before training. Note that the SequentialSampler is utilized to control batches after samples are reordered.

\begin{lstlisting}
## (1) At epoch start, run forward pass to get representations X, Y in the paired dataset. 
model.eval()
with torch.no_grad():
    X, Y = [], []
    for batch in train_dataloader:
        X.append(model(inputs=batch[0]))
        Y.append(model(inputs=batch[1]))

    ## (2) Compute an approx to permutation which minimizes bandwidth of \pi XY^T \pi^T for entries greater than quantile q.
    permutation = compute_perm_bandwidth_min(X, Y, quantile=q)

    ## (3) Reorder the dataset on the approximate solution.
    train_dataset = torch.utils.data.Subset(train_dataset, permutation)
    train_sampler = SequentialSampler(train_dataset)
    train_dataloader = DataLoader(train_dataset, 
                                 sampler=train_sampler, 
                                 batch_size=train_batch_size)

model.train()
## (4) Continue training.
\end{lstlisting}

\section{Dataset Details}

In Table \ref{sent-embed-datasets} and Table \ref{code-search-datasets}, we provide details for all Sentence Embedding and Code Search datasets respectively.

\begin{table}[!ht]
\small
\centering
\begin{tabular}{lc|c|c} \toprule
    Setting & Name & \# of samples  & Source \\ \midrule
    Train & SNLI+MNLI (entailment+hard neg) & 275,602  & \href{https://huggingface.co/datasets/princeton-nlp/datasets-for-simcse/resolve/main/nli_for_simcse.csv}{Hugging Face Download}  \\
    \hline
    Test & STS12 & 3.1K & \href{https://huggingface.co/datasets/princeton-nlp/datasets-for-simcse/resolve/main/senteval.tar}{Hugging Face Download}  \\ 
    Test & STS13 & 1.5K  & \href{https://huggingface.co/datasets/princeton-nlp/datasets-for-simcse/resolve/main/senteval.tar}{Hugging Face Download} \\
    Test & STS14 & 3.7K & \href{https://huggingface.co/datasets/princeton-nlp/datasets-for-simcse/resolve/main/senteval.tar}{Hugging Face Download}  \\
    Test & STS15 & 8.5K &  \href{https://huggingface.co/datasets/princeton-nlp/datasets-for-simcse/resolve/main/senteval.tar}{Hugging Face Download} \\
    Test & STS16 & 9.2K &  \href{https://huggingface.co/datasets/princeton-nlp/datasets-for-simcse/resolve/main/senteval.tar}{Hugging Face Download} \\
    Test & STS-B & 1.4K &  \href{https://huggingface.co/datasets/princeton-nlp/datasets-for-simcse/resolve/main/senteval.tar}{Hugging Face Download} \\
    Test & SICK-R & 4.9K &  \href{https://huggingface.co/datasets/princeton-nlp/datasets-for-simcse/resolve/main/senteval.tar}{Hugging Face Download}  \\
    \bottomrule
\end{tabular}
\caption{Description of training and evaluation datasets for sentence embedding tasks, all datasets are from \cite{gao2021simcse} and further details can be found in the repository.}
\label{sent-embed-datasets}
\end{table}

\begin{table}[!ht]
\small
\centering
\begin{tabular}{lc|c|c|c|c} \toprule
    Name & Train samples & Validation &  Test Samples & \# of Candidates & Source \\ \midrule
    CosQA & 20,000 & 604 & 1,046 & 1,046 & \href{https://github.com/microsoft/CodeBERT/tree/master/UniXcoder/downstream-tasks/code-search#1-advtest-dataset}{CodeBERT Repo}  \\
    AdvTest & 251,820 & 9,604 & 19,210 & 19,210 & \href{https://github.com/microsoft/CodeBERT/tree/master/UniXcoder/downstream-tasks/code-search#2-cosqa-dataset}{CodeBERT Repo} \\
    CSN Go & 167,288 & 7,325 & 8,122 & 28,120 & \href{https://github.com/microsoft/CodeBERT/tree/master/UniXcoder/downstream-tasks/code-search#3-csn-dataset}{CodeBERT Repo}  \\
    CSN Java & 164,923 & 5,183 & 10,955 & 40,347 & \href{https://github.com/microsoft/CodeBERT/tree/master/UniXcoder/downstream-tasks/code-search#3-csn-dataset}{CodeBERT Repo}  \\
    CSN JavaScript & 58,025 & 3,885 & 3,291 & 13,981 & \href{https://github.com/microsoft/CodeBERT/tree/master/UniXcoder/downstream-tasks/code-search#3-csn-dataset}{CodeBERT Repo}  \\
    CSN PHP & 241,241 & 12,982 & 14,014 & 52,660 & \href{https://github.com/microsoft/CodeBERT/tree/master/UniXcoder/downstream-tasks/code-search#3-csn-dataset}{CodeBERT Repo} \\
    CSN Python & 251,820 & 13,914 & 14,918 & 43,827 & \href{https://github.com/microsoft/CodeBERT/tree/master/UniXcoder/downstream-tasks/code-search#3-csn-dataset}{CodeBERT Repo}  \\
    CSN Ruby & 24,927 & 1,400 & 1,261 & 4,360 & \href{https://github.com/microsoft/CodeBERT/tree/master/UniXcoder/downstream-tasks/code-search#3-csn-dataset}{CodeBERT Repo}  \\
    \bottomrule
\end{tabular}
\caption{Description of training and evaluation datasets for code search tasks, all datasets are from \cite{feng-etal-2020-codebert} and further details can be found in the repository.}
\label{code-search-datasets}
\end{table}

\section{Hyperparameters}
\label{Hparams}

In Tables \ref{hparams-sent-embed} and \ref{hparams-code-search} below, we detail the hyperparameters used for the best performing sentence embedding and code search models respectively.

\begin{table}[!ht]
\centering
\begin{tabular}{lc|c|c|c} \toprule
    {Model} & Learning Rate & Batch Size  & Number Epochs & Quantile $q$  \\ \midrule
    SimCSE BERT$_{base}$ & $3e{-5}$ & 256 & 5 & 0.999  \\
    SimCSE RoBERTa$_{base}$ & $3e{-5}$ & 256 & 5 & 0.999  \\
    SimCSE RoBERTa$_{large}$ & $7e{-6}$ & 256 & 5 & 0.9999  \\
    \hline
    PromCSE BERT$_{base}$ & $7e{-3}$ & 256 & 10 & 0.999  \\
    PromCSE RoBERTa$_{base}$ & $7e{-3}$ & 256 & 10 & 0.999  \\
    PromCSE RoBERTa$_{large}$ & $7e{-3}$ & 256 & 10 & 0.999  \\
    \bottomrule
\end{tabular}
\caption{Hyperparameters for best experimental results in Sentence Embedding tasks. }
\label{hparams-sent-embed}
\end{table}

\begin{table}[!ht]
\centering
\begin{tabular}{lc|c|c|c} \toprule
    {Task} & Learning Rate & Batch Size  & Number Epochs & Quantile $q$  \\ \midrule
    CosQA & $2e{-5}$ & 64 & 10 & 0.999  \\
    AdvTest & $2e{-5}$ & 64 & 10 & 0.999  \\
    CSN Ruby & $2e{-5}$ & 64 & 10 & 0.999  \\
    CSN Go & $2e{-5}$ & 64 & 10 & 0.999  \\
    CSN JS & $2e{-5}$ & 64 & 10 & 0.999  \\
    CSN Python & $2e{-5}$ & 64 & 10 & 0.999  \\
    CSN Java & $2e{-5}$ & 64 & 10 & 0.999  \\
    CSN PHP & $2e{-5}$ & 64 & 10 & 0.999  \\
    \bottomrule
\end{tabular}
\caption{Hyperparameters for best experimental results in Code Search tasks for the UniXcoder model. }
\label{hparams-code-search}
\end{table}

For Sentence Embedding tasks, hyperparameters do not vary significantly, other than the learning rate, between models and are similar to those used in the original models with random sampling \cite{gao2021simcse, jiang2022promcse}. For Code Search tasks, we do not vary hyperparameters from the default values from the original paper using random sampling \cite{guo-etal-2022-unixcoder} and, as a result, we use identical settings to the UniXcoder paper other than batch assignments.

\section{Comparison of batch loss values between GCBS, Random Sampling, and Hard Negative Mining}

\begin{figure}[!ht]
\centering
\includegraphics[width=0.79\textwidth]{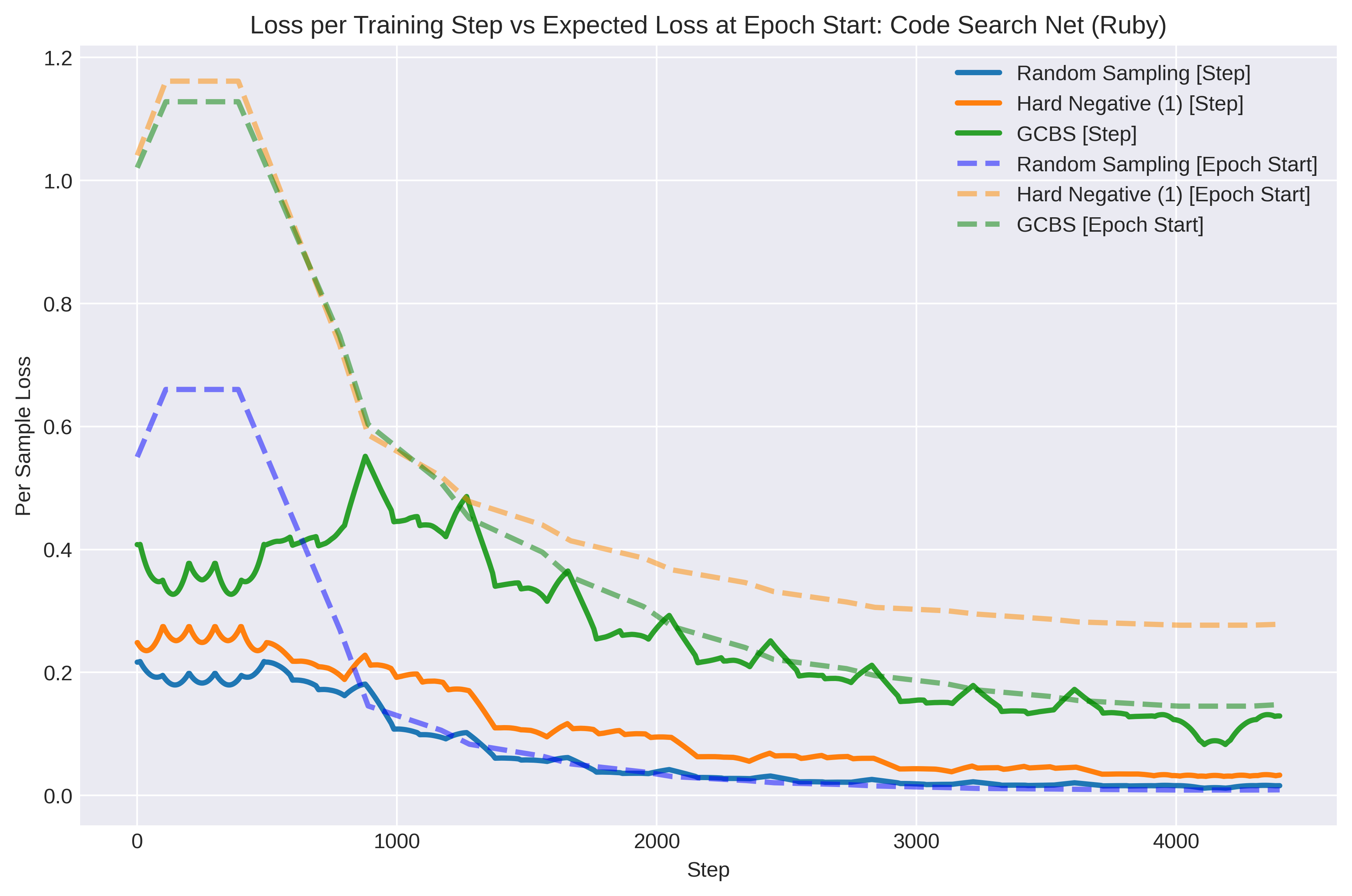}
\caption{Loss per sample and expected loss computed at epoch start for GCBS, Random Sampling, and Hard Negative (1) for the Code Search Net (Ruby) dataset with the UniXcoder model.}
\label{loss_step_2}
\end{figure}

In Figure \ref{loss_step_2}, we show the loss per sample vs step number in the Code Search Net (Ruby) dataset with the UniXCoder model for Random Sampling, GCBS, and mining $1$ hard negative per sample at the beginning of each epoch which we denote as \textit{Hard Negative (1)}. This hard mining approach has a smaller computational burden compared with approaches commonly used in practice but incurs 2x the runtime of GCBS and 3x the runtime of random sampling as detailed in Section \ref{sec:runtimes}. 

Additionally, we compare each training step loss to the expected loss over batches calculated at the beginning of each epoch. This requires performing a forward pass at the start of each epoch, assigning batches, and then computing the loss over in-batch negatives for each sample. After the first few epochs, while the expected loss over batches for Random Sampling and GCBS is well approximated by the expected loss at the epoch start, expected losses for Hard Negative Mining are substantially overestimated. This corroborates findings in previous literature \cite{Wang2021CrossBatchNS, xiong2021approximate} which motivates the need to update nearest neighbor indices frequently within an epoch, further increasing the computational burden of Hard Negative Mining. Empirically, we find that the observed loss per sample for GCBS is significantly larger than that of Random Sampling or Hard Negative (1) and, like Random Sampling but not Hard Negative (1), can be well approximated by the expected loss at the epoch start. Losses are smoothed as a running average over the previous $500$ training steps.

\section{Runtime scaling of GCBS vs number of samples $N$}

In order to empirically characterize the runtime scaling of GCBS with respect to the number of samples $N$, we run simulations with random embeddings of dimension 768 both in 1 GPU and 7 GPU settings. Results for these experiments are included in Table \ref{Runtime_scale_1} and Table \ref{Runtime_scale_7}. For all experimentation, we scale the quantile value to keep 512 expected values in each row/column after sparsification (i.e. $q = 1 - \frac{512}{N}$).

\begin{table}[h]
\centering
\begin{minipage}{0.48\linewidth}
\centering
\begin{tabular}{l|c} \toprule
    {N} & Runtime (seconds) \\ \midrule
    10000 & 2.56 \\
    50000 & 14.77 \\
    100000 & 34.68 \\
    500000 & 322.52 \\
    1000000 & 1013.19 \\
    \bottomrule
\end{tabular}
\caption{Runtime scaling of GCBS on a single A40 GPU with respect to number of samples $N$}
\label{Runtime_scale_1}
\end{minipage}
\begin{minipage}{0.48\linewidth}
\centering
\begin{tabular}{l|c} \toprule
    {N} & Runtime (seconds) \\ \midrule
    10000 & 31.05 \\
    50000 & 38.17 \\
    100000 & 52.85 \\
    500000 & 241.91 \\
    1000000 & 591.55 \\
    5000000 & 3956.39 \\
    \bottomrule
\end{tabular}
\caption{Runtime scaling of GCBS on 7 A40 GPUs with respect to number of samples $N$}
\label{Runtime_scale_7}
\end{minipage}
\label{tab:models}
\end{table}

We find that on a single A40 GPU, $N = 10^6$ is tractable in under 20 minutes and for 7 GPUs $N=5 \times 10^6$ is tractable in under 1 hour. We believe for very large scale datasets, initial space partitioning with a nearest neighbors library (i.e. FAISS) and then using our approach on partitions of size $10^6$ may be beneficial. An exact implementation of profiled code in these experiments for 1 GPU experiments are provided below.

\begin{lstlisting}
import torch
import timeit
import math  
from scipy.sparse.csgraph import reverse_cuthill_mckee    
from scipy.sparse import csr_matrix
import statistics

def compute_gcbs(z1_outs, z2_outs, quantile):                                                                                                             
      start_time = timeit.default_timer()
      # (1) Stack and normalize outputs             
      src_train_full = torch.nn.functional.normalize(z1_outs).cuda()
      tgt_train_full = torch.nn.functional.normalize(z2_outs).cuda()
      z1_outs, z2_outs = [], []
      
      # (2) Estimate quantile
      chunk_size, num_samples, quantiles = 3, len(tgt_train_full), []
      for chunk_idx in range(math.ceil(len(tgt_train_full)/chunk_size)):
        mat_val = src_train_full[chunk_idx*chunk_size:(chunk_idx+1)*chunk_size] @ tgt_train_full.T
        quantiles.append(float(torch.quantile(mat_val, quantile)))

      # (3) Get similarity graph thresholded on quantile
      row, col, data, quantile = [], [], [], statistics.median(quantiles)
      for chunk_idx in range(math.ceil(len(src_train_full)/chunk_size)):
        mat_val = src_train_full[chunk_idx*chunk_size:(chunk_idx+1)*chunk_size] @ tgt_train_full.T
        ret = (mat_val.flatten() > quantile).nonzero(as_tuple=True)[0].cpu()
        row += ((ret - (ret % num_samples))/num_samples + chunk_idx*chunk_size).int().tolist()
        col += (ret % num_samples).tolist()
        data += [1.0 for _ in range(len(ret))]

      # (4) Get permutation using graph bandwidth minimization on sparsified graph (cuthill-mckee)  
      permutation = list(reverse_cuthill_mckee(csr_matrix((data, (row, col)),
                                                    shape=(num_samples, num_samples)))) 
      print(timeit.default_timer() - start_time)
      return permutation

for size in [10000, 50000, 100000, 500000, 1000000]:
    keep_per_sample = 512
    z1, z2 = torch.rand(size, 768), torch.rand(size, 768)
    quantile = 1 - float(keep_per_sample/size)
    perm = compute_gcbs(z1, z2, quantile)
\end{lstlisting}

\section{Variation in Performance across Random Seeds}

Since our proposed method GCBS is a deterministic algorithm, the only randomness in performance is from the parameter initialization of the pooling layer. We conduct 5 runs of the experiments with different seeds. The standard deviations on sentence embedding tasks with BERT base are:

\begin{table*}[!ht]
\tiny
\centering
\begin{tabular}{lc|c|c|c|c|c|c|c} \toprule
    {Model} & STS12 & STS13  & STS14 & STS15 & STS16 & STS-B & SICK-R & Avg  \\ \midrule
    SimCSE BERT$_{base}$ w/ GCBS & 75.82 $\pm$ 0.06 & 85.30 $\pm$ 0.02 & 81.12 $\pm$ 0.17 & 86.58 $\pm$ 0.11 & 81.68 $\pm$ 0.05 & 84.80 $\pm$ 0.01 & 80.04 $\pm$ 0.05 & 82.19 $\pm$ 0.05  \\
    \bottomrule
\end{tabular}
\caption{The performance and standard deviations across seeds for SimCSE BERT$_{base}$ \cite{gao2021simcse} with GCBS. The reported score is Spearman correlation magnified by a factor of 100.}
\label{table:sent-embed-variation}
\end{table*}

The standard deviation on the CosQA code search task with the UniXcoder model are:

\begin{table*}[!ht]
\centering
\begin{tabular}{lc} \toprule
    {Model} & CosQA \\ \midrule
    UniXcoder w/ GCBS & 71.1 $\pm$ 0.26  \\
    \bottomrule
\end{tabular}
\caption{The performance and standard deviations across seeds for the UniXcoder \cite{guo-etal-2022-unixcoder} model with GCBS. The reported score is Mean Reciprical Rank magnified by a factor of 100.}
\label{table:code-embed-variation}
\end{table*}

The standard deviation of GCBS's average performance on sentence embedding tasks is 0.05\%, while the the relative performance is 0.62\%. The standard deviation of GCBS's performance on CosQA is 0.26\%, while the relative improvement is 1.0\%.

\section{Limitations}
Our GCBS approach, while efficient compared to alternatives, does increase the runtime of standard contrastive learning approaches (i.e. SimCLR, Moco). As a result, it increases training costs for experiments which would be limiting for some researchers. In large scale settings (i.e. $> 5 \times 10^6$ paired samples), additional steps to partition the training data for parallel processing may be required before using our approach.

\end{document}